# Receptivity of an AI Cognitive Assistant by the Radiology Community: A Report on Data Collected at RSNA

Karina Kanjaria[1][a], Anup Pillai[1][b], Chaitanya Shivade[2][c], Marina Bendersky[3], Ashutosh Jadhav[1],
Vandana Mukherjee[1][d], and Tanveer Syeda-Mahmood[1]

[1]*Almaden Research Center, IBM, Harry Rd, San Jose, USA*
[2]*Amazon Web Services, Amazon, University Ave, Palo Alto, USA*
[3]*Data Science, Nevro, Bridge Pkwy, Redwood City, USA*
{kanjaria, apillai, vandana, stf}@us.ibm.com, shivade.1@buckeyemail.osu.edu, marina.bendersky@gmail.com

Abstract: Due to advances in machine learning and artificial intelligence (AI), a new role is emerging for machines as intelligent assistants to radiologists in their clinical workflows. But what systematic clinical thought processes are these machines using? Are they similar enough to those of radiologists to be trusted as assistants? A live demonstration of such a technology was conducted at the 2016 Scientific Assembly and Annual Meeting of the Radiological Society of North America (RSNA). The demonstration was presented in the form of a question-answering system that took a radiology multiple choice question and a medical image as inputs. The AI system then demonstrated a cognitive workflow, involving text analysis, image analysis, and reasoning, to process the question and generate the most probable answer. A post demonstration survey was made available to the participants who experienced the demo and tested the question answering system. Of the reported 54,037 meeting registrants, 2,927 visited the demonstration booth, 1,991 experienced the demo, and 1,025 completed a post-demonstration survey. In this paper, the methodology of the survey is shown and a summary of its results are presented. The results of the survey show a very high level of receptiveness to cognitive computing technology and artificial intelligence among radiologists.

## 1 INTRODUCTION

Radiologists and other clinicians who interpret or otherwise employ medical imaging in their practices are increasingly overwhelmed by the explosion of images and associated data. While picture archiving and communication systems and other electronic health record systems may help organize, store, and present such data (Zeleznik et al., 1983), the burden of data overload contributes to physician burnout and medical errors (Singh et al., 2013). Computing technology, including the use of artificial intelligence and deep learning, could provide a solution by providing caregivers with cognitive assistance. Examples of potential roles for this technology include: triaging images to exclude those that are certainly normal from clinician review, finding abnormalities to draw attention to images or exams that are certainly or likely abnormal, intelligently selecting prior exams or images for comparison, methodically registering comparison images, systematically tracking index lesions, compiling and presenting clinical information based on the indications of a particular exam, presenting differential diagnoses with associated probabilities based on clinical and imaging findings, segmenting normal and abnormal anatomy on medical images, protecting reading physicians against known cognitive biases, and helping generate the clinical report. When considering any of these roles for artificial intelligence, some in the medical imaging community have raised concerns regarding the future role of physicians and the threat of their replacement by cognitive computing. Past studies have shown that although decision support systems improve the workflow of a clinician, healthcare decision makers took longer to acknowledge their value and the resulting benefits (Pynoo et al., 2012). To better understand these concerns, a demonstration using an artificial intelligence technology was conducted at RSNA 2016. The participants were subsequently surveyed to understand receptivity to the project and the demonstrated system, what features of the technology were important to them, and to gather meaningful opinions, comments, and feedback.

---

[a] https://orcid.org/0000-0002-6964-6286
[b] https://orcid.org/0000-0002-7130-2909
[c] https://orcid.org/0000-0001-6604-1129
[d] https://orcid.org/0000-0002-8189-328X

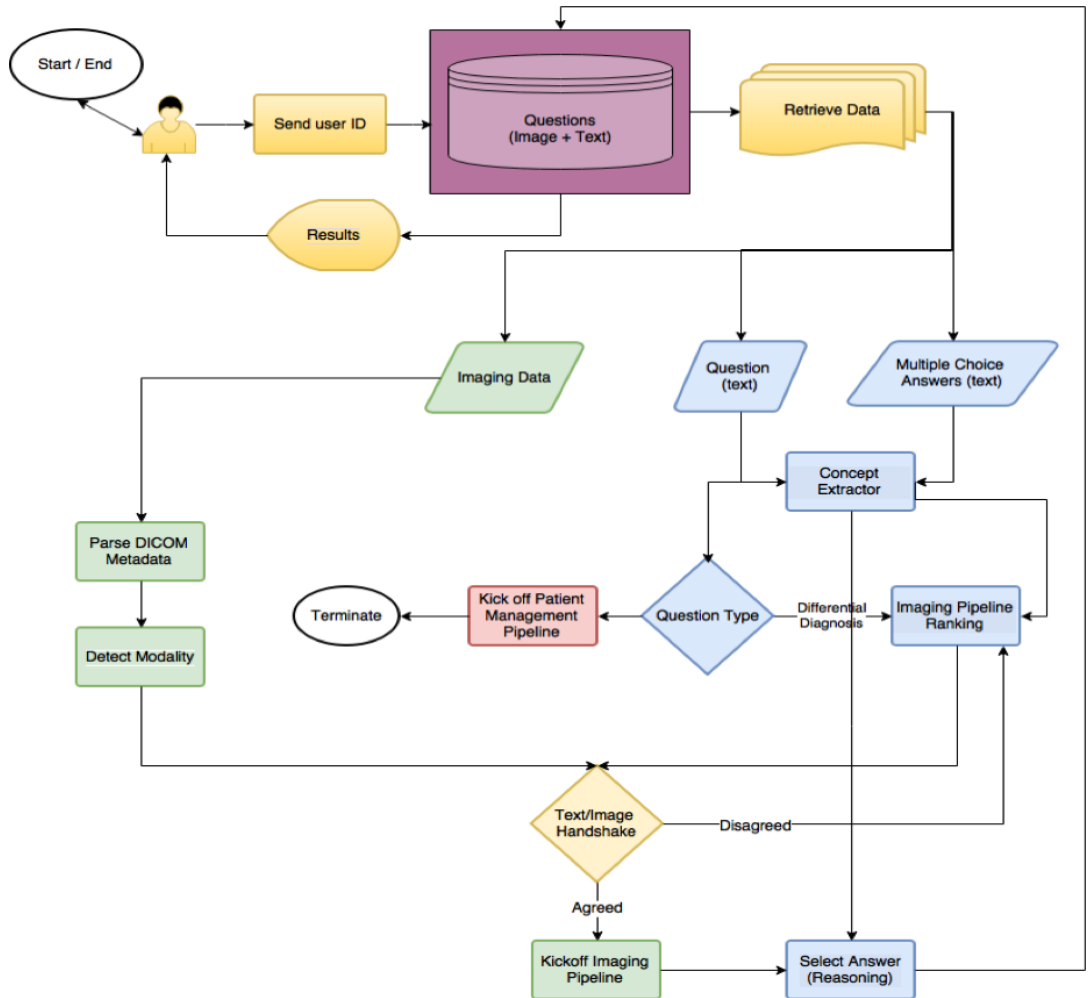

Figure 1: Architecture diagram for the Eyes of Watson system.

## 2 SYSTEM ARCHITECTURE

The AI system was designed, specifically, as a Question Answering (QA) system that attempted to answer radiology questions formulated in natural language. We focused on two main question types: i) differential diagnosis and ii) patient management. An answer could be i) an anomaly for differential diagnosis questions or ii) a procedure for management type questions. In addition, for every answer suggested by the system, a justification explaining how the system utilized the available textual and imaging evidence to arrive at an answer was also presented using Sankey charts (Google, 2016). The system leveraged multiple disease detection algorithms processed on clinical images and knowledge extracted from the textual information presented, to arrive at answers weighted by probabilities based on both the imaging and textual evidence. The system highlighted the search and extraction of evidence from our medical knowledge database, built by mining resources such as medical ontologies, textual corpora and clinical websites.

Some such mined resources were Unified Medical Language System (UMLS), SNOMED CT, Radlex, Dynamed, and Diagnosis Pro.

A detailed overview of our QA system is shown in **Figure 1**. When a user logged in to the system, they were presented with multiple studies to choose from. Upon selection, they were shown a question with a brief patient history, multiple answer choices, and an imaging study related to the question. After they made their answer selection, they could have chosen to ask the system to analyze the study. This began the analysis process in which the text and imaging data were, first, extracted separately and sent to the appropriated processing stages. Following this, the textual and imaging information underwent a sanity-check stage, which served as a verification process to corroborate that the information gathered was in sync. Finally, this information was forwarded to the reasoning stage which performed answer ranking based on evidence, and presented the answers weighted by their probabilities.

The main segments of our system are elaborated below and further details may be found in (Pillai et al., 2019).

### 2.1 Question Classification

In order for a QA system to answer a question with speed and efficiency, the system must first be able to deduce the type of question it is presented with. This question type classification becomes even more difficult than traditional text classification problems when the question text contains only a few short sentences.

To address these complications with question classification, we employed the use of a Naïve Bayes text classifier. The classifier applied Bayes theorem with independent feature selection and could be applied more easily to larger sets of data. It was simpler than other models, which aided in real-time classification tasks (Rish et al., 2001). Our classifier made the assumption that the presence of a feature in a class is unrelated to the presence of another feature in the class. The classifier also assumed that numerous features selected together created a higher probability of the question being classified in the correct category- "differential diagnosis" versus "patient management".

Differential diagnosis questions are those which require an analysis on a list of symptoms and information to come to a conclusive finding based on the evidence. For example, the following question would be classified as a differential diagnosis question: "A 61-year-old female presents to the ER with a heart murmur and acute chest pain. Past medical history is positive for acute myocardial infarction. Patient also has an extensive history of tobacco use. Given the following chest X-ray, which is most likely?".

Patient management problems also require an analysis of the facts presented, however, they focus on the next steps of treatment. The following would be classified as a patient management problem: "A 52-year-old male with hypertension is admitted to the Emergency Department with acute chest pain, tachypnea, back pain. Unenhanced CTA was performed and it shows intimal calcification toward the lumen. A contrast CTA was performed next and is shown here. What is the next step of management?".

The Naïve Bayes classifier we chose suited best for the real-time reasoning application we aimed for, because of its ability to speedily predict question types in real-time during the demonstration. It did this by counting the number of times a word appears amongst all of the words in a question with a specific label, and then assigning a label to the question after looking at its array of words and its class. After question text classification into differential diagnosis or patient management, the system then proceeded to text and image processing.

### 2.2 Text and Image Processing

The text and image processing stage is further delineated into four major steps as follows.

i) Concept and feature identification. In this step, we identified the most relevant concepts and features in the presented data. These included, but were not limited to, things such as demographic information, signs and symptoms, medications, image type, treatments, and question type, along with negated concepts. The concepts were extracted by running a concept extractor on the question text, and only the most relevant information was retained by the reasoning module.

ii) Feature expansion. In this step, we expanded the identified concepts and features with their synonyms, related words, and ontological concepts. We did this in order to address the mismatch that derived from extracted concepts in questions differing from those in our knowledge base. For example, a question text could contain the words "heart attack" while our knowledge base refers to this as "myocardial infarction". Since these two concepts are related, we needed to perform feature expansion to ensure that they were mapped together.

iii) Knowledge retrieval. In this step, we fetched all the relevant clinical knowledge for every possible disease based on the expanded features. Our knowledge base consists of scientific facts that capture disease and diagnosis information in the forms of con-cepts and relationships. For example, a relationship in its simplest form in the knowledge base can be represented plainly as "chest pain: sign and symptom: myocardial infarction".

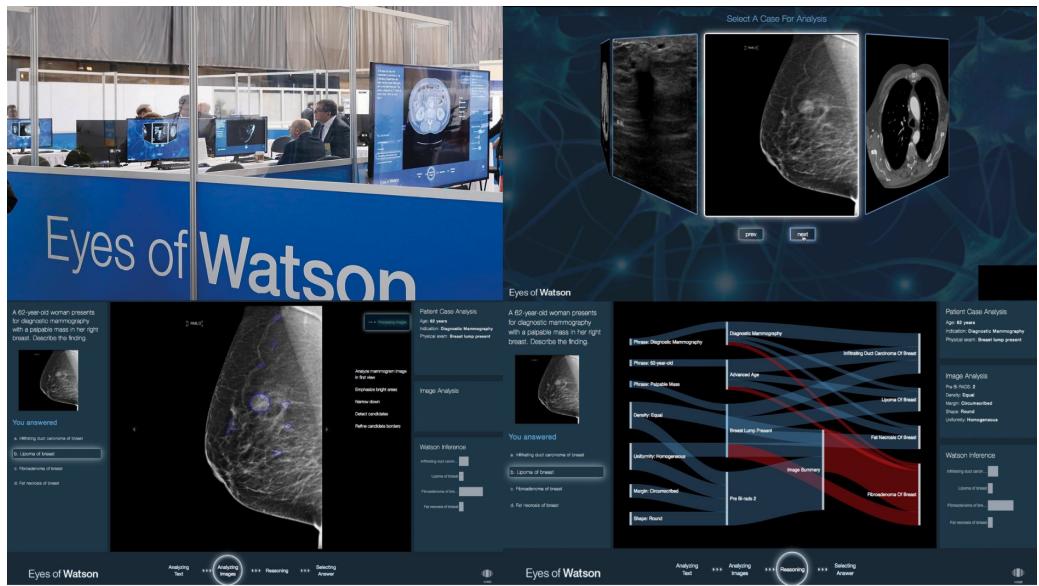

Figure 2: The Eyes of Watson exhibit at RSNA. Users received brief instructions and were provided access to a workstation (upper left), could select exams (upper right), were presented images with a multiple-choice question (lower left), and could then opt to view the results and the clinical inference pathway in a Sankey chart produced by the system (lower right).

iv) Imaging pipeline ranking. In this step, we ranked the answer choices based on all of the previous steps in conjunction with the question type information. This module determined which imaging pipeline was the most pertinent to the possible answer choices. An analytics framework then registered this information. The framework contained all imaging pipelines and some related textual information (e.g. pipeline descriptions and imaging modalities). These features were inputs to the search for applicable imaging pipelines within the analytics framework. If Digital Imaging and Communications in Medicine (DICOM) header information was available, this information was also included as a feature in the search. The candidate pipeline search results were filtered into further stages. These stages involved duplicates removal and updated pipeline selection. In order to differentiate between analogous pipelines, we performed distance calculations between the text descriptions and concepts extracted for each pipeline and answer. The pipelines were, finally, ranked in accordance with the shortest distance. For example, in the case of two similar pipelines, "aortic dissection" and "aortic aneurysm" such filtering became invaluable and essential. Once all of the answer choices were matched with imaging pipelines, the pipelines were run in sequential order.

Our answer selection process was an iterative one. The features gathered from this text and image processing module were fed to the reasoning module after the highest ranked imaging pipeline performed its analysis. Had the reasoning module not been able to provide sufficient evidence to probabilistically support an answer, the next highest ranked imaging module was run. In this way, the process continued to iterate until sufficient confidence in an answer was acquired. Our reasoning module and how it was visualized by the system is detailed next.

## 2.3 Reasoning and Visualization

From each imaging pipeline, we identified features such as diseases, specific measurements, and imaging abnormalities. We then retained evidence from both the imaging features and the textual features. From this information, the reasoning algorithm was able to create a probabilistic ranking of the answer choices/targeted diseases. The disease or answer choice with the highest probability was then selected based on this evidence presented.

In our system, we chose to display our reasoning workflow in the form of Sankey charts (Riehmann et al., 2005). They are used to visualize what steps the reasoning algorithm performed in order to arrive at its conclusion. An example of such a Sankey chart can be seen in **Figure 3**. The chart shows how the most probable answer choice from the acquired data points was determined by displaying data flows in the form of weighted and colored lines (Google, 2016). A line's thickness denotes the weight of that data point and its underlying confidence. Different colors are also used to represent categories and show transitions between stages. Ultimately, the highest ranked answer choice would present with the highest number and largest thickness of lines, showcasing that the particular answer had the most supporting evidence in quantity and quality.

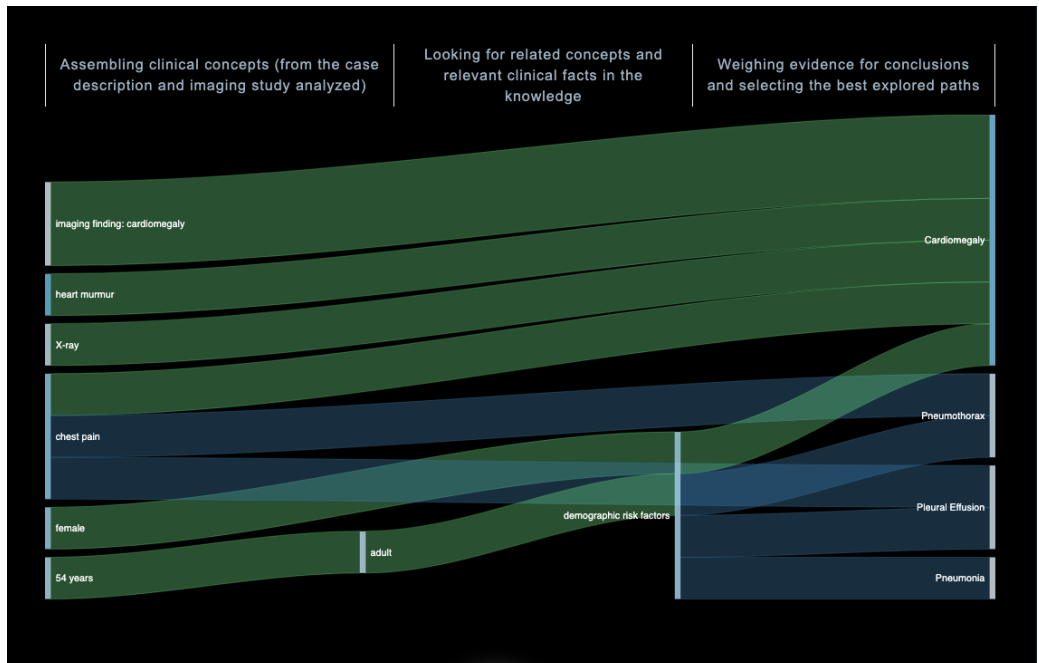

Figure 3: Example Sankey chart used to visualize the system's reasoning steps.

## 3 MATERIALS AND METHODS

### 3.1 Setting

The system's exhibit was assembled in the learning center at RSNA 2016 in Chicago. This conference was attended by 54,037 registrants, including 26,988 healthcare professionals (RSNA, 2017).

A team of researchers and engineers arranged for a computer demonstration wherein attendees could use one of 16 kiosks that were provided. Three types of conference attendees visited the demonstration at the kiosks: i) A person who simply visited the learning center, but did not sign in to a kiosk with their credentials or experience the demonstration first-hand; ii) A person who visited the learning center, signed in to the kiosk with their credentials, experienced the demonstration, but did not take the survey; iii) A person who visited the learning center, signed in to the kiosk, experienced the demonstration, and took the survey.

Visitors belonging to the first category were mostly passive onlookers who experienced the demonstration second-hand at a kiosk where someone else was signed in. Those who visited the learning center were first provided two training videos, each 2.5 minutes in duration. Persons who were interested in learning more and experiencing the demonstration, then progressed to a kiosk. Each kiosk had four random cases selected from a collection of approximately 200 mammography and cardiovascular cases. The collection included images in the form of chest computed tomography (CT), chest computed tomography angiography (CTA), cardiac magnetic resonance imaging (MRI), and digital radiographs. Each of the cases were prepared in a question-answering format by a team of radiologists who worked with the engineers, with the intent to mimic real world cases.

In each case, the user was presented with the images along with a multiple-choice question with four possible answer choices. The user had the option to choose an answer and then click "Ask Watson" to show the most probable answer, or they could opt out of choosing an answer and directly click "Ask Watson" to show the workflow. In either case, by asking the system, the user could experience and visualize how our cognitive computing technology solved the same exam and answered the same question. The cognitive process was deliberately slowed down so that the participants could see the system analyze the question text, parse out the salient clinical information, and then evaluate the images and segment the normal anatomy. In the next step, the system evaluated the images for suspected pathology. During each step of the process, as more data were collected and analyzed, the AI system calculated and updated the statistical probability of each answer choice. In the final step, the thought process of the system was graphically illustrated through the Sankey diagrams. Following the completion of the process, the participant could return to each step to peruse. Using the clinical inference pathway page, participants could hover over each color coded pathway to see the weight the system gave that data point, and how it was connected to the possible answer choices.

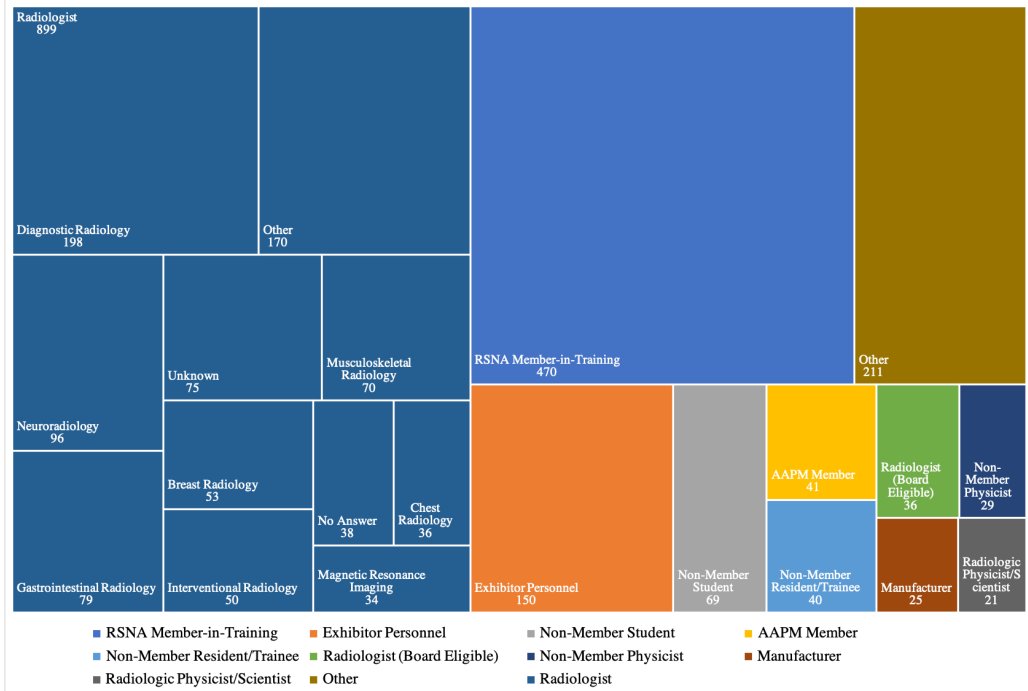

Figure 4: The 1991 participants in the system demonstration outlined by type, including radiology speciality.

After experiencing the demonstration, users were provided with a questionnaire. Their responses were then tabulated and reported herein.

## 3.2 Data Analysis

There were 2,927 total participants who experienced the Eyes of Watson demonstration. Described as visitor type one above, these individuals visited the exhibit in the learning center and scanned their RSNA conference badge to indicate their presence. A total of 1,991 of these visitors proceeded to sign in to a kiosk and take an exam (visitor types two and three). This indicates that most, 67%, of the visitors to the learning center were interested in learning more about the proposed system upon seeing the kiosks and training videos.

Furthermore, we analyzed the type of user who partook in the demonstration, and found that 45% were radiologists, while the rest were RSNA members, exhibitor personnel, or other types of individuals. We also analyzed the type of radiologist specializations, in order to better understand whether the system would be of interest to experts outside of the chest and breast radiology fields, since our system was focused in those two radiology specialties. We found that a variety of radiologists specializing in different fields participated in the exams; chest and breast ra-diologists accounted for fewer than 100 participants. The types of users depicted by their titles, and the further detailed radiologist speciality information is shown in **Figure 4**. The figure also includes the number of each type of participant.

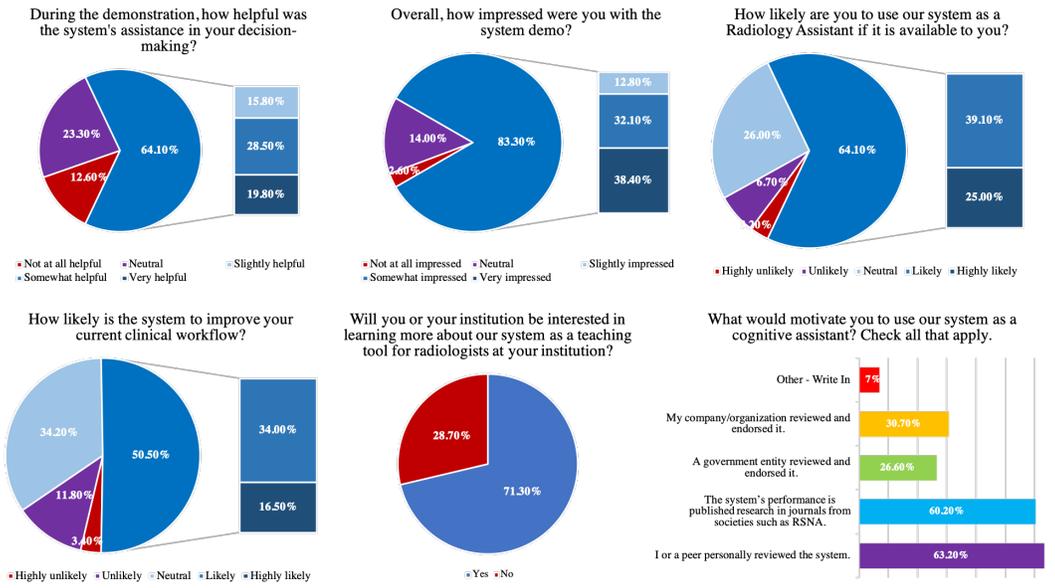

Figure 5: Survey multiple choice questions and responses. Top row and bottom left questions use Likert scale. Bottom middle is a "yes or no" question. Bottom right question allows for multiple selections.

Visitors who took an exam spent an average of 43 seconds per question in the demo, and, overall, users responded to a total of 7,023 questions. In 4,235 instances, the participant both answered a question and selected the "Ask Watson" option. This option allowed the system to perform its cognitive process on the question. In 70 instances, the participant only answered the question. In 2,129 instances, the user asked the system without first answering the question. Lastly, in 589 instances, the user neither answered the question nor asked the system. When analyzed, this shows that around 91% of users were interested in seeing how the system worked once they were exposed to it. Additionally, on average, each participant answered less than one question correctly. This suggests that users were not motivated by the idea of solving the problem correctly or outperforming the system. Verbal feedback collected by staff at the event indicated that many users purposefully chose an incorrect answer in order to test the system. To support this anecdotal evidence, data collected showed that 51% of users answered the questions in less than 30 seconds- an insufficient amount of time to arrive at the correct conclusion. Analysis showed that users spent an average of 41 seconds per question when the question was answered incorrectly. In contrast, users took an average of 51 seconds per question to choose the correct answer. This shows that spending more time on the questions led to correct answer selection, however, most users chose to spend 30 seconds or less.

Lastly, we provided a questionnaire with seven questions for users who experienced the demonstration first-hand. This questionnaire was curated by our team as a means to understand users' experience with our system. We included questions with Likert scale instead of simply polar questions in order to better understand the degree of receptiveness. A total of 1,025 users participated in the survey and answered all of the questions following the demonstration; these users were defined as visitor type three above. Six of the questions possessed a multiple choice format, while the remaining one involved a free text response. Of the six multiple choice questions, four questions employed the use of the Likert scale, one was a "yes or no" question, and one consisted of custom answer choices designed by the authors. The questionnaire, which can be visualized in **Appendix A**, was administered electronically using a standard survey tool (Surveymonkey, 2019). **Figure 5** shows the user response to the multiple choice portion of the questionnaire. All positive responses such as "slightly impressed", "somewhat impressed", and "very impressed" were grouped together to better showcase the true receptivity to the system. The percent distinctions between these categories can also be seen in the figure. Ad-ditionally, **Appendix B** presents some representative responses to the free text question that show the most popular feedback.

## 4 SUMMARY OF SURVEY RESULTS

The survey results can be summarized as follows: Overall, the RSNA attendees who received a demonstration of the AI tool were impressed with the system and receptive to adoption. Nearly 85% were impressed by the technology, and the majority reported that they would like to use such a tool in the future. 15% of survey takers did not believe that a cognitive assistant would help improve their workflow, exhibiting some skepticism. As seen in **Figure 4**, the majority of participants also indicated that they would use such a tool if validated by their own or a peer's personal experience, or if shown to be valid in a peer-reviewed publication. Respondents, reportedly, were less likely to be convinced by a government entity or local validation.

Despite the much-published concern that AI systems may threaten physician jobs, the RSNA attendees' responses suggest that they are most motivated by clinical outcomes. As AI continues to improve, it will likely become a regularly used tool in all aspects of healthcare. An intelligent cognitive assistant could be a major factor that helps reduce clinical workload and allows physicians to focus on their primary purpose – the patient. Our results indicate that most people who participated in this experience are open and ready for a future of AI augmented medicine.

While this study showed promise for AI aided cognitive assistants, it has a few limitations. First, we did not rigorously collect accuracy data or validate the credentials of the test-takers; our information was derived from data collected from scanned RSNA badges and system interaction. As a result, we cannot report whether the user's impressions were influenced by their individual performance relative to the system. Their receptivity could conceivably be influenced by a technology that is either so outstanding that it is threatening, or so poor that it is not helpful. Secondly, we could not control against bias. For example, we do not know if the attendees were self-selected because of their interest in this technology. We did not collect data that might have indicated whether the exhibit changed a user's impression of the technology.

Nevertheless, the results indicate that imaging professionals are open to the use of artificial intelligence technologies to provide cognitive assistance, particularly if validated by personal experience, a peer reference, or published research.

## APPENDIX A

**Question 1.** During the demonstration, how helpful was the system's assistance in your decision-making?

```
a. Not at all helpful
b. Slightly helpful
c. Neutral
d. Somewhat helpful
e. Very helpful
```

**Question 2.** Overall, how impressed were you with the system demo?

```
a. Not at all impressed
b. Slightly impressed
c. Neutral
d. Somewhat impressed
e. Very impressed
```

**Question 3.** How likely are you to use our system as a Radiology Assistant if it is available to you?

```
a. Highly unlikely
b. Unlikely
c. Neutral
d. Likely
e. Highly Likely
```

**Question 4.** How likely is the system to improve your current clinical workflow?

```
a. Highly unlikely
b. Unlikely
c. Neutral
d. Likely
e. Highly Likely
```

**Question 5.** Will you or your institution be interested in learning more about using our system as a teaching tool for radiologists at your institution?

```
a. Yes
b. No
```

**Question 6.** What would motivate you to use our system as a cognitive assistant? Check all that apply.

```
a. I or a peer personally reviewed the
system's performance.
b. The system's performance is published
research in journals from societies such
as RSNA.
c. A government entity reviewed and
endorsed it.
d. My company/organization reviewed and
endorsed it.
e. Other
```

**Question 7.** Where do you think this should be included in your clinical workflow? (Do you have any additional feedback?)

```
Answer: Free text
```

# APPENDIX B

**Sample of free text responses to Question 7:**

| |
|---|
| During routine interpretation At emergency room by non-radiologists |
| It would be helpful if it reviewed findings after I have written the report but before I have signed it off, allowing me to re-review my findings. |
| As head of a large academic radiology department training more residents than any other program in the US, I think this could first be educational for residents, and would be worth exploring in breast and PE imaging. |
| Pre-reading cases Peer review Help with difficult cases Synthesize clinical data in context with imaging |
| Should be integrating into the reading workflow. Watson could be inserted between the time of imaging acquisition and image interpretation by the radiologist. |
| 1. Every exam should be evaluated and a report (within limits) be generated and then compared to the final read. For example, outpatient studies that sit on the list over the weekend or institutions that do not have 24 hours coverage. The read generated by Watson should not be available for clinical decision making until properly endorsed by an accredited body or institution. 2. Post-operative changes, however, a standardized post-operative note should be made available for a simpler algorithm. 3. Long nodule follow-up 4. Any patient with cross-sectional imaging that requires follow-up. 5. HCC screening!!! |
| in case of doubts: we use to ask colleagues for suggestions in difficult cases, but on night calls or in small hospitals it may happen you have no colleagues whom to ask for. therefor, my suggestion is it should not be to expansive so small hospital ans small clinical practicians may be buy it |
| People will usually choose the cheapest option. Machine is going to be cheaper than human. Thus performance must be absolutely verified before clinical implementation because it can not, I believe, be put back into the box. Also, I listened a lecture about AI and I would suggest AI as a second reader (if there are two readers to all studies), because it would be more rewarding for a radiologist to do the job first and then only check the potential misses. If radiologists job is only to check if the machine has made any mistakes, I believe A) she/he would more easily ignore machine's suggestions and B)she/he becomes demoralized. (Sloppy, underperfoming, unenthusiastic and depressed). |